\documentclass[conference]{IEEEtran}
\IEEEoverridecommandlockouts
\usepackage{cite}
\usepackage{amsmath,amssymb,amsfonts}
\usepackage{algorithmic}
\usepackage{graphicx}
\usepackage{textcomp}
\usepackage{xcolor}
\usepackage{amsmath,amsfonts}
\usepackage{algorithmic}
\usepackage{algorithm}
\usepackage[caption=false,font=scriptsize,labelfont=sf]{subfig}
\usepackage{babel}
\usepackage{array}
\usepackage{textcomp}
\usepackage{stfloats}
\usepackage{booktabs}
\usepackage[export]{adjustbox}
\usepackage{url}
\usepackage{verbatim}
\usepackage{graphicx}
\usepackage{cite}
\usepackage{bm}
\usepackage{accents}
\usepackage{array}
\usepackage{multicol}
\usepackage{multirow}
\usepackage{eqparbox}
\usepackage{enumerate}
\usepackage{nccmath}
\usepackage{makecell}

\newcolumntype{M}[1]{>{\centering\arraybackslash}m{#1}}
\newcolumntype{P}[1]{>{\centering\arraybackslash}p{#1}}

\def\BibTeX{{\rm B\kern-.05em{\sc i\kern-.025em b}\kern-.08em
    T\kern-.1667em\lower.7ex\hbox{E}\kern-.125emX}}
    
\begin{document}

\title{Optimal Energy Storage Scheduling for Wind Curtailment Reduction and Energy Arbitrage: \\ A Deep Reinforcement Learning Approach 
\thanks{*Corresponding author: Hao Wang.}
\thanks{This work was supported in part by the Australian Research Council (ARC) Discovery Early Career Researcher Award (DECRA) under Grant DE230100046.}
}

\author{\IEEEauthorblockN{Jinhao Li\textsuperscript{1}, Changlong Wang\textsuperscript{2,3}, Hao Wang\textsuperscript{1,3*}}
\IEEEauthorblockA{\textsuperscript{1}Department of Data Science and AI, Faculty of IT, Monash University, Australia \\
\textsuperscript{2}Department of Civil Engineering, Monash University, Australia\\
\textsuperscript{3}Monash Energy Institute, Monash University, Australia\\
Email: stephlee175@gmail.com, chang.wang@monash.edu, hao.wang2@monash.edu
}
}

\maketitle

\begin{abstract}
Wind energy has been rapidly gaining popularity as a means for combating climate change. However, the variable nature of wind generation can undermine system reliability and lead to wind curtailment, causing substantial economic losses to wind power producers.
Battery energy storage systems (BESS) that serve as onsite backup sources are among the solutions to mitigate wind curtailment. However, such an auxiliary role of the BESS might severely weaken its economic viability. This paper addresses the issue by proposing joint wind curtailment reduction and energy arbitrage for the BESS.
We decouple the market participation of the co-located wind-battery system and develop a joint-bidding framework for the wind farm and BESS. It is challenging to optimize the joint-bidding because of the stochasticity of energy prices and wind generation. Therefore, we leverage deep reinforcement learning to maximize the overall revenue from the spot market while unlocking the BESS's potential in concurrently reducing wind curtailment and conducting energy arbitrage. We validate the proposed strategy using realistic wind farm data and demonstrate that our joint-bidding strategy responds better to wind curtailment and generates higher revenues than the optimization-based benchmark. Our simulations also reveal that the extra wind generation used to be curtailed can be an effective power source to charge the BESS, resulting in additional financial returns. 
\end{abstract}
% \vspace*{-1em}
\begin{IEEEkeywords}
Deep reinforcement learning, energy arbitrage, spot market, wind-battery system, wind curtailment.
\end{IEEEkeywords}

\section{Introduction} \label{sec:intro}
To mitigate climate change and support the global energy transition to net-zero, wind energy has been widely adopted as the main pillar for decarbonization in modern power systems. In 2021, wind contributed $9.9\%$ to Australia's total electricity production, making it the largest utility-scale renewable source~\cite{CEC2022}.
However, the intermittent nature of wind power and inaccurate wind forecasts make it greatly challenging to accommodate wind generation in real-time. Sometimes, wind curtailment is necessary to ensure system security and reliability~\cite{burke2011} at the expense of wind producers. The pace of wind adoption has led to an increase in wind curtailment, which causes considerable losses for wind producers. Similarly, the adoption of battery energy storage systems (BESS) is also gaining momentum: approximately $650$ MW has been registered in the Australian National Electricity Market (NEM), and an additional $34.3$ GW is planned in the next decade~\cite{aemo_future_gen}. 

The integration of renewable energy and BESS is becoming increasingly popular, as the co-location of them can effectively reduce renewable curtailment, diversify revenue streams, mitigate market risks, and defer network augmentation. Co-located renewable energy and BESS system is demonstrated to be effective in the Integrated System Plan~\cite{AEMOSysPlan2022} by the Australian Energy Market Operator (AEMO). 
Given the rapid adoption of co-located wind-battery systems, it is critical to develop effective coordination strategies, not only for the benefit of the power grid but also for enhancing the viability of the BESS as part of a seamless energy transition.

In the co-located wind-battery system, the BESS can act as a storage medium to reduce wind curtailment by absorbing the surplus wind generation. The optimal sizing and scheduling of the BESS were studied in \cite{sun2017,alanazi2017,nikoobahkt2020} via stochastic or robust optimization based on prior knowledge of the wind power uncertainty distribution. Nevertheless, such an auxiliary role may limit the BESS's economic potential, such as performing energy arbitrage (i.e., buy low and sell high), a significant revenue stream for the BESS in the electricity market.

Optimization-based methods have also been used to study wind-battery coordinated bidding strategies in the electricity market. These studies~\cite{akbari2019,khojasteh2021,xie2021} treated the wind farm and the BESS as two independent players to bid in an aggregated manner, while mitigating wind curtailment has been neglected in the arbitrage process of the BESS.
Also, the effectiveness of the proposed strategies highly relies on the accuracy of energy price forecasting. However, accurate energy price prediction is notoriously difficult since the spot market is highly volatile and the price drivers are remarkably complicated \cite{weron2014}. 

Apart from the optimization-based approaches, there has been a lack of research on real-time bidding of the wind-battery coupled system using other avenues. To bridge the literature gap, we develop a novel deep reinforcement learning (DRL)-based bidding strategy for the co-located wind-battery system to concurrently reduce wind curtailment while maximizing the overall revenue through energy arbitrage in the electricity spot market. With its model-free characteristics, the DRL can learn the uncertainties of wind generation and electricity prices from historical observations, i.e., without prior knowledge or price forecasting. Additionally, the online and interactive nature of DRL makes it promising to dynamically balance the trade-off between energy arbitrage and wind curtailment mitigation in real-time bidding for the wind-battery system. 
Our developed joint bidding of the wind-battery system via deep reinforcement learning is referred to as ``JointDRL''.
Our main contributions are summarized as follows.
\begin{itemize}
    \item \emph{Synergizing wind curtailment management and BESS energy arbitrage}: We explore synergies between wind curtailment management and BESS energy arbitrage of a co-located wind-battery system in electricity spot markets. Our research emphasizes the importance of dynamic coordination strategies between renewable generation and storage in achieving profitability.
    \item \emph{DRL-based joint bidding}: We decouple the wind-battery system's market participation into two joint-bidding processes for the wind farm and the BESS. A cutting-edge model-free DRL algorithm, known as twin delayed deep deterministic policy gradient (TD3), is introduced to learn and optimize the joint-bidding strategy.
    \item \emph{Numerical simulations and implications}: We validate our JointDRL in the NEM using realistic wind farm data. Our results demonstrate the effectiveness of our method and reveal the synergy between wind curtailment management and BESS energy arbitrage. The BESS can reduce wind curtailment by charging otherwise curtailed wind power to boost economic returns from energy arbitrage.
\end{itemize}

The rest of paper is organized as follows. Section \ref{sec:model} formulates the participation of the wind-battery system in the electricity spot market. We decouple the bidding process of the wind-battery system and introduce DRL in Section \ref{sec:method} to concurrently maximize the overall revenue and reduce wind curtailment. Simulation results are presented and discussed in Section \ref{sec:exp}, and Section \ref{sec:conclusions} concludes this paper.

\section{System Model} \label{sec:model}
\begin{figure}[!t]
    \centering
    \includegraphics[width=0.8\linewidth]{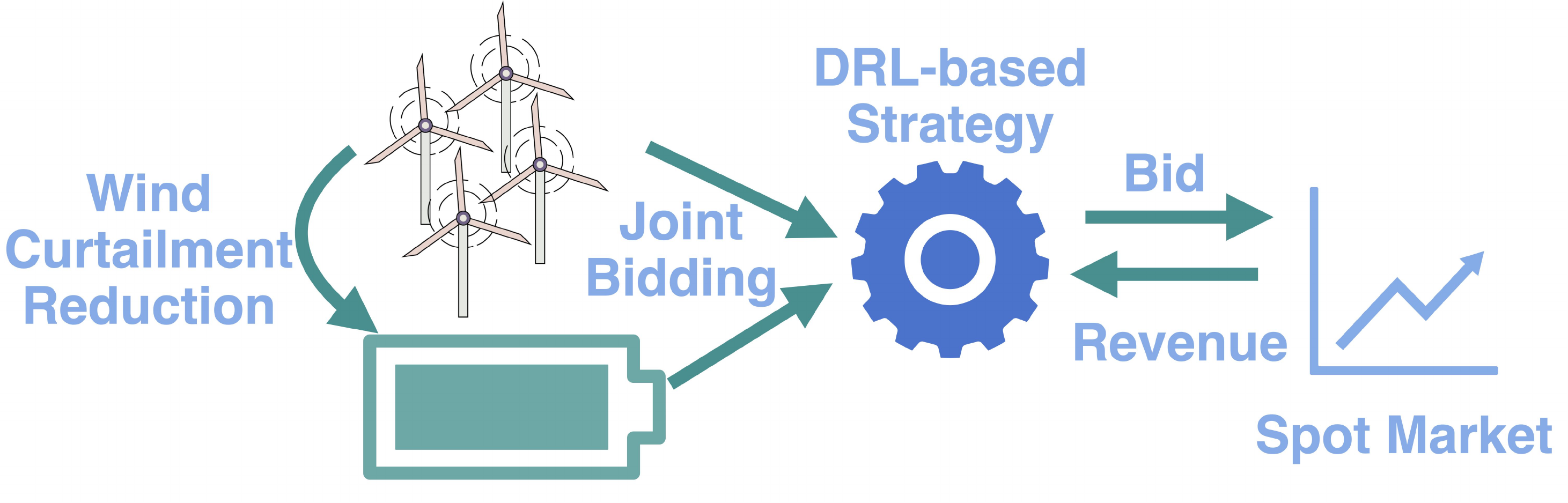}
    \vspace{-0.5em}
    \caption{An overview of the JointDRL bidding strategy.}
    \label{fig:framework}
    \vspace{-1.25em}
\end{figure}

We develop the JointDRL by assuming the wind-battery system is a price taker; thus, its bids will not affect other generator bidding decisions or market clearing outcomes. We also assume the wind farm and the BESS are co-located, and there is sufficient capacity within the substation and transmission lines to allow for concurrent export from both facilities. The BESS dynamically manages onsite wind curtailment while conducting energy arbitrage in the spot market. The context of bidding is discussed in detail in Section \ref{subsec:model_preliminaries}. Section \ref{subsec:model_revenue_gen} outlines the wind and BESS revenue streams under various operational conditions. Section \ref{subsec:model_formulation} formulates the joint bidding of the wind-battery system with wind curtailment management. An overview of the JointDRL is illustrated in Fig. \ref{fig:framework}.

\subsection{The NEM Spot Market} \label{subsec:model_preliminaries}
As part of the NEM, the spot market is a real-time market for trading wholesale electricity between generators and loads, where power supply and demand are balanced instantaneously through a centrally coordinated dispatch process managed by the AEMO~\cite{aemo2020}. Generators submit bids (price and quantity) every five minutes. AEMO dispatches generators in a least-cost manner by ranking generator bids from low to high to form a bidding stack. The generator bids that fulfil the last demand in the bidding stack determine the market clearing price, known as the spot price. Generators that bid below or at that price will get dispatched at their offered quantity and get paid at the clearing price. 

\subsection{Revenue of the Wind-Battery System} \label{subsec:model_revenue_gen}
\subsubsection{Wind Farm} \label{subsubsec:model_revenue_gen_wind}
Wind farms are often registered as semi-dispatchable generators in the NEM and required to constantly update their forecasted generation availability (from onsite wind monitoring devices), denoted by $p_t^\text{W}$, to AEMO, based on which a dispatch target (in MWh) is instructed to the wind farm~\cite{aemo2022} to fulfil in the next dispatch interval. The variable nature of wind power, however, could lead to deviations between the dispatch target and the actual wind generation $p_t^\text{W,Act}$, which subsequently influence the amount of power sent out from the wind farm. We assume the dispatch target will be fully met at times when there is sufficient wind ($p_t^\text{W,Act}>p_t^\text{W}$), whereas the dispatch target can only be partially met ($p_t^\text{W,Act}<p_t^\text{W}$) if there is a wind shortage caused by forecasting errors or non-compliance with market rules~\cite{aemo2022}. We define the final dispatched wind power as $\min\{p_t^\text{W,Act},p_t^\text{W}\}$. We also impose a penalty on the wind farm if it fails to meet the dispatch target. We present the spot market revenue generated by the wind farm, denoted by $R^\text{W}$, as
\begin{equation}
    \label{eq:wind_revenue}
    R^\text{W} = \Delta t\sum_{t=1}^T \rho_t\left( \min\{p_t^\text{W,Act},p_t^\text{W}\}-\lambda|p_t^\text{W,Act}-p_t^\text{W}|\right),
\end{equation}
where $\Delta t$ is the NEM dispatch interval, i.e., $5$ minutes; $T$ is the overall time frame; $\rho_t$ is the spot market clearing price; and $\lambda$ is a penalty coefficient for deviations between the actual wind generation and the AEMO dispatch target~\cite{xie2021}.

\subsubsection{BESS} \label{subsubsec:model_revenue_gen_BESS}
Spot market volatility often motivates the BESS to engage in energy arbitrage. Providing that the BESS cannot simultaneously charge and discharge, we introduce two binary variables $v_t^\text{Ch},v_t^\text{Dch}$ to prevent this from happening, which can be formulated as
\begin{equation}
    \label{eq:cons_BESS_ch_dch}
    v_t^\text{Dch} + v_t^\text{Ch} \leq 1, \quad v_t^\text{Dch},v_t^\text{Ch} \in \{0,1\},
\end{equation}
where the BESS sits idle when these two variables are set to zero.

The BESS's revenue from the spot market, denoted by $R^\text{BESS}$, is shown as
\begin{equation}
    \label{eq:BESS_revenue}
    R^\text{BESS} = \Delta t\sum_{t=1}^T\left(v_t^\text{Dch}\eta^\text{Dch}-v_t^\text{Ch}\frac{1}{\eta^\text{Ch}}\right)\rho_t p_t^\text{BESS,S},
\end{equation}
where $\eta^\text{Ch}$/$\eta^\text{Dch}$ are charging/discharging efficiencies of the BESS and $p_t^\text{BESS,S}$ is the bid power in the spot market. Apart from buying power in the spot market, the otherwise curtailed wind generation can also be a potential power source to charge the BESS. We denote the power planned to draw from the onsite wind farm as $p_t^\text{BESS,WC}$. Following the charging/discharging constraint in Eq. \eqref{eq:cons_BESS_ch_dch}, the BESS cannot charge itself using onsite wind curtailment when bidding to discharge in the spot market, which can be formulated in logic as
\begin{equation}
    \label{eq:cons_BESS_WC}
    v_t^\text{Dch}p_t^\text{BESS,WC} = 0.
\end{equation}

Also, frequent charging/discharging lead to cycle aging of the BESS. We define the battery degradation cost $C^\text{BESS}$ as
\begin{equation}
    \label{eq:BESS_degradation_cost}
    C^\text{BESS} = c\Delta t\sum_{t=1}^T v_t^\text{Dch}p_t^\text{BESS,S},
\end{equation}
where we approximate the battery degradation as a result of discharging~\cite{bordin2017,anwar2022}; $c$ is a specific battery technology cost-coefficient in AU\$/MWh~\cite{anwar2022}.

\subsection{Joint Bidding of the Wind-Battery System} \label{subsec:model_formulation}
Considering the distinct revenue streams of the wind farm and the BESS from the spot market, along with the degradation cost of the BESS, we formulate the optimal joint bidding of the wind-battery system as an optimization problem whose objective is expressed as
\begin{equation}
    \label{eq:objective}
    \max \hspace{0.25em} R^\text{W} + R^\text{BESS} - C^\text{BESS}.
\end{equation}

Real-time dispatch of the wind farm and BESS are constrained by
\begin{align}
    \label{eq:cons_wind_power}
    0&\leq p_t^\text{W}\leq P_\text{max}^\text{W},\\
    \label{eq:cons_BESS_spot_power}
    0&\leq p_t^\text{BESS,S}\leq P_\text{max}^\text{BESS},\\
    \label{eq:cons_BESS_WC_power}
    0&\leq p_t^\text{BESS,WC}\leq P_\text{max}^\text{BESS},\\
    \label{eq:cons_BESS_tot_power}
    0&\leq p_t^\text{BESS,S}+p_t^\text{BESS,WC}\leq P_\text{max}^\text{BESS},
\end{align}
where $P_\text{max}^\text{W}$ and $P_\text{max}^\text{BESS}$ are the installed capacity (in MW) of the wind farm and the rated power (in MW) of the BESS, respectively. Eq. \eqref{eq:cons_wind_power} constrains the forecasted availability of the wind farm. Eq. \eqref{eq:cons_BESS_spot_power} and \eqref{eq:cons_BESS_WC_power} represent the power that the BESS can bid in the spot market and the power planned to draw from the wind farm must be within the rated power of the BESS. Furthermore, Eq. \eqref{eq:cons_BESS_tot_power} shows that the sum of the bid and power drawn from the onsite wind farm cannot exceed the rated power of the BESS.

Also, charging/discharging operations of the BESS are limited by its current capacity $e_{t-1} + \Delta e_t$, where $e_{t-1}$ is its capacity after the previous dispatch interval, and $\Delta e_t$ is the energy change in the current dispatch interval. The BESS's capacity must be within its lower and upper energy limits denoted by $E_\text{min}$ and $E_\text{max}$, which can be formulated as
\begin{equation}
    \label{eq:cons_BESS_energy_change}
    E_\text{min} \leq e_{t-1} + \Delta e_t\leq E_\text{max}.
\end{equation}
The BESS's capacity fluctuates due to power exchange in the spot market or drawing the otherwise curtailed energy from the onsite wind farm. We define curtailed wind power as
\begin{equation}
    \label{eq:wind_curtailment}
    p_t^\text{W,WC} = \left(p_t^\text{W,Act}-p_t^\text{W}\right)\mathbb{I}\left(p_t^\text{W,Act}>p_t^\text{W}\right),
\end{equation}
where $\mathbb{I}\left(p_t^\text{W,Act}>p_t^\text{W}\right)$ is an indicator of wind curtailment. Thus, the energy change $\Delta e_t$ in Eq. \eqref{eq:cons_BESS_energy_change} can be written as
\begin{equation}
\label{eq:def_BESS_energy_change}
\Delta e_t = \Delta t\left[\left(v_t^\text{Ch}-v_t^\text{Dch}\right)p_t^\text{BESS,S}+\min\left\{p_t^\text{BESS,WC},p_t^\text{W,WC}\right\}\right],
\end{equation}
where the first term represents the energy change from bidding and the second term indicates the BESS's response to wind curtailment.

\section{Methodology} \label{sec:method}
To maximize the overall revenue of the wind-battery system formulated, we decouple the continuous bidding problem into two Markov decision processes (MDP) for the wind farm and the BESS in Section \ref{subsec:method_MDP}, followed by Section \ref{subsec:method_TD3}, where TD3~\cite{fujimoto2018} is introduced to maximize the expected returns of the derived MDPs, which facilitates the optimization of the entire revenue-oriented bidding problem.

\subsection{MDP Modeling} \label{subsec:method_MDP}
As discussed, multiple factors can affect how the wind-battery coupled system bids in the spot market. Joint bidding can be better characterized by decoupling it into two MDPs (for the wind farm and the BESS, respectively), each with four elements:
$\mathbb{S}^\text{W}$/$\mathbb{S}^\text{BESS}$, $\mathbb{A}^\text{W}$/$\mathbb{A}^\text{BESS}$, $\mathbb{P}^\text{W}$/$\mathbb{P}^\text{BESS}$, and $\mathbb{R}^\text{W}$/$\mathbb{R}^\text{BESS}$.

\textbf{State Space} $\mathbb{S}$: All internal (e.g., wind generation) and external (e.g., energy prices) information can be represented as a state $\bm{s}_t$. To guide the BESS's response to wind curtailment, we introduce wind curtailment frequency within the latest $L$ dispatch intervals, denoted by $f_{t}^\text{WC}$, in the state of the BESS. States of the wind farm and the BESS are defined as
\begin{equation}
    \label{eq:MDP_state}
    \hspace{-0.3em}\bm{s}_t^\text{W} = \left[p_{t-1}^\text{W,Act},\rho_{t-1}\right],
    \bm{s}_t^\text{BESS} = \left[e_{t-1},f_{t-1}^\text{WC},p_{t-1}^\text{W,Act},\rho_{t-1}\right].
\end{equation}

\textbf{Action space} $\mathbb{A}$: Action of the wind farm represents its forecasted availability $p_t^\text{W}$, while actions of the BESS are bid power $p_t^\text{BESS,S}$ and power drawn from the onsite wind curtailment $p_t^\text{BESS,WC}$. Actions of the wind farm and the BESS are normalized to range from $0$ to $1$ and formulated as
\begin{equation}
    \label{eq:MDP_action}
    \bm{a}_t^\text{W} = \left[a_t^\text{W}\right],
    \hspace{0.25em}
    \bm{a}_t^\text{BESS} = \left[v_t^\text{Dch},v_t^\text{Ch},a_t^\text{BESS,S},a_t^\text{BESS,WC}\right],
\end{equation}

\textbf{Probability space} $\mathbb{P}$: The probability space refers to the probability set of transitioning to the next state after taking a deterministic action, which is defined as $\mathbb{P}\left(\bm{s}_{t+1}|\bm{s}_t,\bm{a}_t\right)$.

\textbf{Reward Space} $\mathbb{R}$: The wind farm and BESS receive rewards after taking action $\bm{a}_t$ at state $\bm{s}_t$, which reflects the effectiveness of the bidding decision. To monitor wind generation uncertainty and update accurate dispatch targets, we formulate the reward function of the wind farm $r_t^\text{W}$ as
\begin{equation}
    \label{eq:wind_tot_reward}
    r_t^\text{W} = -\rho_t|a_t^\text{W}P_\text{max}^\text{W}-p_t^\text{W,Act}|.
\end{equation}

Effective BESS energy arbitrage is enabled by the introduction of two charging/discharging indicators, denoted by $\mathbb{I}^\text{Ch}_t$/$\mathbb{I}^\text{Dch}_t$, and formulated as
\begin{equation}
    \label{eq:reward_indicator}
    \mathbb{I}_t^\text{Ch} = \text{sgn}\left(\bar{\rho}_t-\rho_t\right),\quad \mathbb{I}_t^\text{Dch} = \text{sgn}\left(\rho_t-\bar{\rho}_t\right),
\end{equation}
where $\text{sgn}(\cdot)$ is the sign function and $\bar{\rho}_t$ is the exponential moving average of the spot price, which is defined as
\begin{equation}
    \label{eq:MCP_moving_avg}
    \bar{\rho}_t = \tau\bar{\rho}_{t-1} + \left(1-\tau\right)\rho_t,
\end{equation}
where $\tau$ is a smoothing parameter. The charging/discharging indicators incentivize the BESS to buy low ($\rho_t<\bar{\rho}_t$) and sell high ($\rho_t>\bar{\rho}_t$). Any bids violating such a guideline will be penalized. The arbitrage reward $r_t^\text{BESS,S}$ is thus formulated as
\begin{fleqn}
\begin{equation}
    \label{eq:BESS_ES_reward}
    r_t^\text{BESS,S} = a_t^\text{BESS,S} |\rho_t - \bar{\rho}_t| \left(\mathbb{I}_t^\text{Ch}v_t^\text{Ch}\frac{1}{\eta^\text{Ch}} + \mathbb{I}_t^\text{Dch}v_t^\text{Dch}\eta^\text{Dch}\right).
\end{equation}
\end{fleqn}

The BESS receives positive rewards, denoted by $r_t^\text{BESS,WC}$, when reducing onsite wind curtailment, formulated as
\begin{equation}
    r_t^\text{BESS,WC} = \beta \min\left\{a_t^\text{BESS,WC},\frac{p_t^\text{W,WC}}{P_\text{max}^\text{W}}\right\} f_t^\text{WC}\frac{1}{\eta^\text{Ch}},
\end{equation}
where $\beta$ is the incentive factor for wind curtailment reduction and the minimum term represents the normalized absorbed wind power. 

The BESS reward function $r_t^\text{BESS}$ combines bidding rewards and wind curtailment mitigation rewards, formulated as
\begin{equation}
    \label{eq_BESS_tot_reward}
    r^\text{BESS}_t = r_t^\text{BESS,S} + r_t^\text{BESS,WC}.
\end{equation}

\subsection{Learning Optimal Bidding Strategy via TD3}
\label{subsec:method_TD3}
We introduce a state-of-the-art off-policy DRL algorithm, referred to TD3~\cite{fujimoto2018}, to optimize the derived MDPs where the same TD3 structure is adopted. TD3 aims to learn an optimal action strategy, denoted by $\pi$, that maximizes the expected returns over a finite horizon, which can be formulated as
\begin{equation}
    \label{eq:DRL_obj}
    J_\pi = \mathbb{E}_{\bm{s}_t\sim \mathbb{P},\bm{a}_t\sim \pi(\bm{s}_t)} \left[\sum_{t=1}^T \gamma^{t-1} r_t\right],
\end{equation}
where $\gamma$ is the discounted factor.

\section{Experiments and Results} \label{sec:exp}
\subsection{Experimental Settings} \label{subsec:exp_setting}
The wind generation data is collected from the Oaklands Hill Wind Farm in Victoria, one of the five jurisdictions of the NEM in Australia. We use Victoria spot prices in $2018$~\cite{aemo_data} to train and evaluate our JointDRL, where the first eleven months are for training and the last month for evaluation. The storage capacity of the BESS is assumed to be $10$ MWh with its minimum and maximum allowable state of charge being $5\%$ and $95\%$, respectively. We used $1$ Nvidia TITAN RTX graphics processing units for algorithm training. The initialized parameters are provided in Table \ref{tab:parameters}.
\begin{table}[!t]
    \centering
    \caption{Initialized parameters.}
    %\vspace{-0.4em}
    \begin{tabular}{cc || cc || cc}
    \hline
    $\Delta t$ & $5$ mins & $\lambda$ & $1.5$ &
    $\eta^\text{Ch},\eta^\text{Dch}$ & $0.95$ \\
    $c$ & $1$ AU\$/MWh & $P_\text{max}^\text{W}$ & $67$ MW & $P_\text{max}^\text{BESS}$ & $10$ MW \\
    $E_\text{min}$ & $0.5$ MWh & $E_\text{max}$ & $9.5$ MWh & 
    $L$ & $10$\\
    $\tau$ & $0.9$ & $\beta$ & $10$ & $\gamma$ & $0.99$\\
    \hline
    \end{tabular}
    \label{tab:parameters}
    \vspace{-1.0em}
\end{table}

\begin{table}[!t]
    \centering
    \caption{The evaluation results of the P\&O and JointDRL.}
    \vspace{-0.5em}
    \begin{tabular}{|c|c|c|c|}
    \cline{3-4}
    \multicolumn{2}{c|}{} & P\&O & \textbf{JointDRL}\\
    \hline
    \multirow{3}{*}{Revenue (AU\$)} & Wind & $829,604$ & $\bm{979,784}$\\
    \cline{2-4}
    & BESS & $62,444$ & $\bm{113,292}$ \\
    \cline{2-4}
    & Total & $892,048$ & $\bm{1,093,076}$ \\
    \hline
    \multirow{2}{*}{\makecell{Wind Curtailment (MWh)}} & Curtailed & $493$ & $\bm{124}$\\
    \cline{2-4}
    & Absorbed & $130$ & $\bm{313}$\\
    % \multicolumn{2}{|c|}{\makecell{Wind Curtailment (MWh)}} & $1/629$ & $\bm{313/437}$\\
    % MW 1*(60/5)=12 629*(60/5)=7536; 313*(60/5)=3756 437*(60/5)=5244
    \hline
    \multirow{3}{*}{\makecell{Execution Time (mins)}} & Train & $2.1$ & $56.2$ \\
    \cline{2-4}
    & Evaluate & $32.2$ & $\bm{0.1}$ \\
    \cline{2-4}
    & Total & $34.3$ & $56.3$ \\
    \hline
    \end{tabular}
    \label{tab:baseline_comparison}
    \vspace{-2.0em}
\end{table}

\subsection{Effectiveness of the JointDRL}
\begin{figure}[!t]
    \centering
    \subfloat[Revenue.]{
    \vspace{-1em}
    \includegraphics[width=0.47\linewidth]{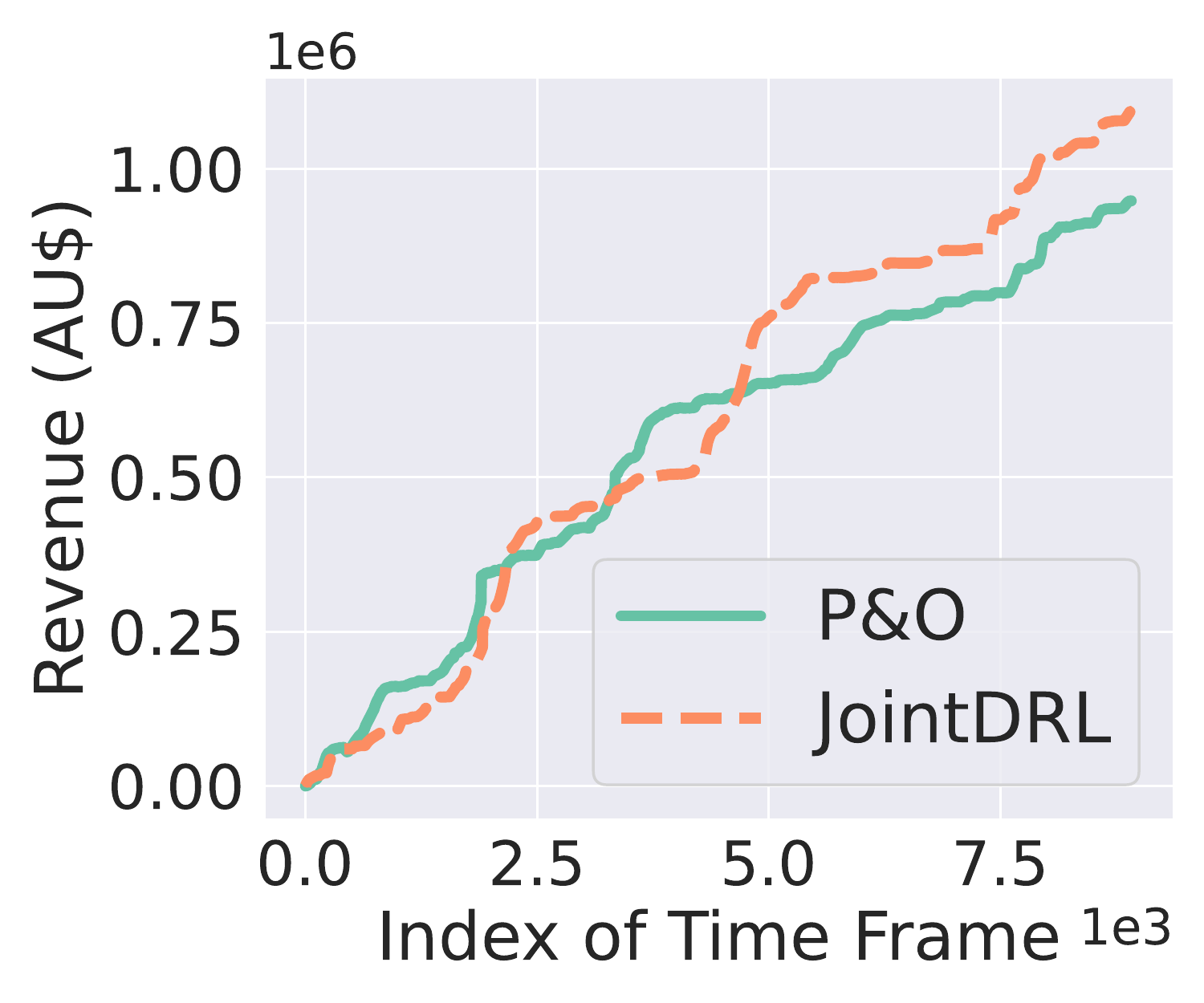}
    \label{fig:baseline_comparison_revenue}
    }
    \subfloat[BESS's power source.]{
    \includegraphics[width=0.47\linewidth]{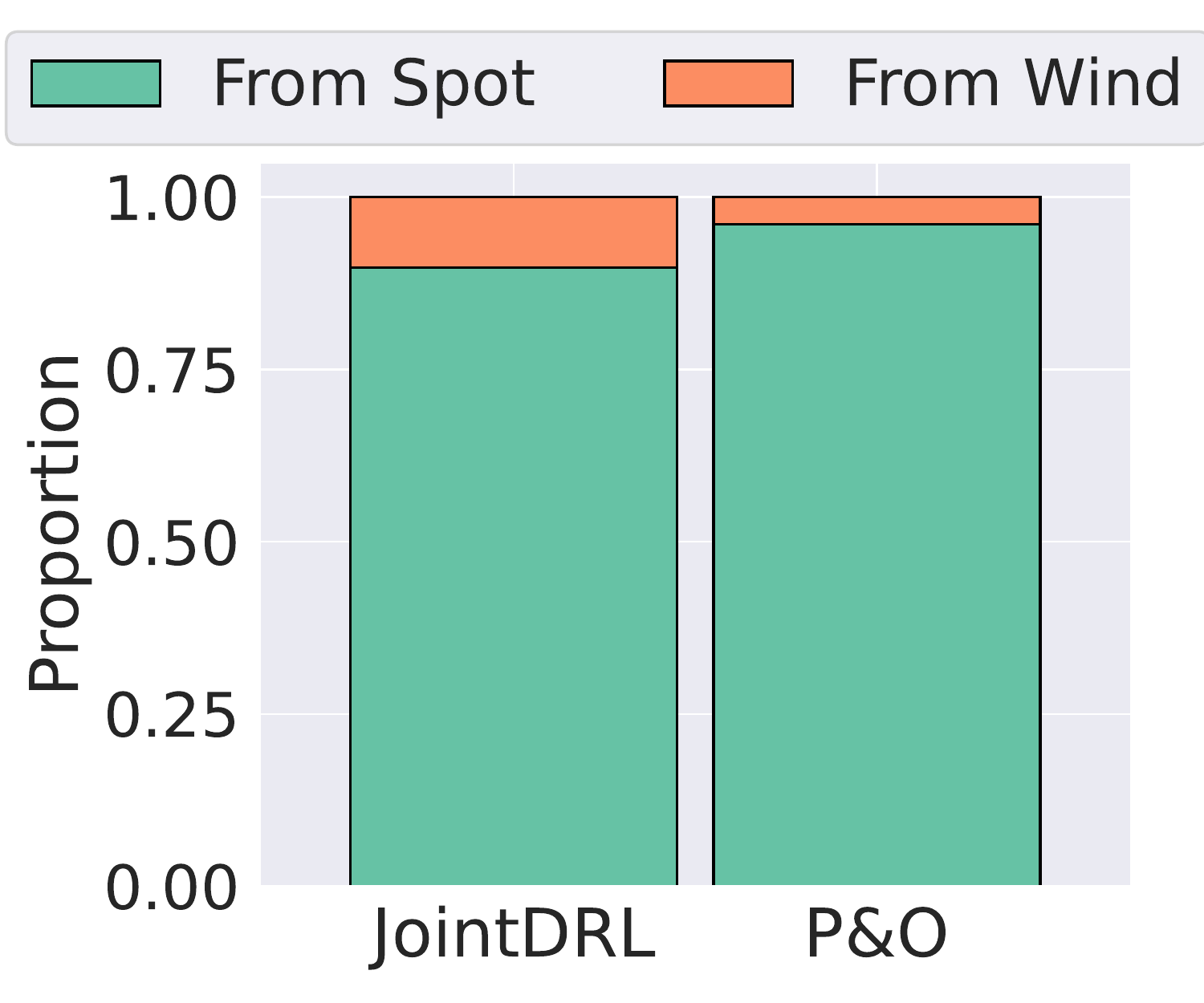}
    \label{fig:baseline_comparison_power_source}
    }
    %\vspace{-0.5em}
    \caption{Performance comparison between JointDRL and the P\&O benchmark.}
    \vspace{-1.5em}
    \label{fig:baseline_comparison}
\end{figure}

To examine the effectiveness of our JointDRL, we develop a predict-and-optimize (P\&O) benchmark for comparison. The P\&O method relies on a long short-term memory (LSTM) network for wind availability and energy price forecasts, and solves the revenue maximization problem by a mixed integer linear programming solver from the PuLP library~\cite{mitchell2011}. The cumulative revenue derived from each method is illustrated in Fig. \ref{fig:baseline_comparison_revenue} with associated statistics presented in Table \ref{tab:baseline_comparison} for cross comparison. The results show the JointDRL outperforms the P\&O benchmark significantly. In particular, our method generates $81\%$ additional revenue for the BESS and $18\%$ for the wind farm compared to the P\&O benchmark. 

From the revenue breakdown in Table~\ref{tab:baseline_comparison}, our JointDRL takes advantage of the onsite wind surplus and performs significantly better in terms of utilizing the curtailed energy, i.e., absorbing $313$ MWh of curtailed wind energy compared to that of $130$ MWh using the P\&O method. The JointDRL's capability in managing wind curtailment avoids a considerable amount of excessive wind generation used to be curtailed. In contrast, the otherwise curtailed wind energy using P\&O is nearly four times higher than ours, as shown in Table \ref{tab:baseline_comparison}. In our case, about $10$\% of the BESS's stored energy comes from curtailed wind energy, as shown in Fig. \ref{fig:baseline_comparison_power_source}.

Despite the fact that the JointDRL requires a longer training time, as shown in Table \ref{tab:baseline_comparison}, it can make bidding decisions at a significantly faster rate, i.e., $10$ seconds for one-month bidding. Therefore, a well-trained JointDRL is better suited to real-time bidding, as accurate and rapid decision-making is essential.

\subsection{Wind Curtailment Reduction} \label{subsec:exp_WPC_reduction}
\begin{figure}[!t]
    \centering
    \subfloat[Revenue.]{
    \includegraphics[width=0.43\linewidth]{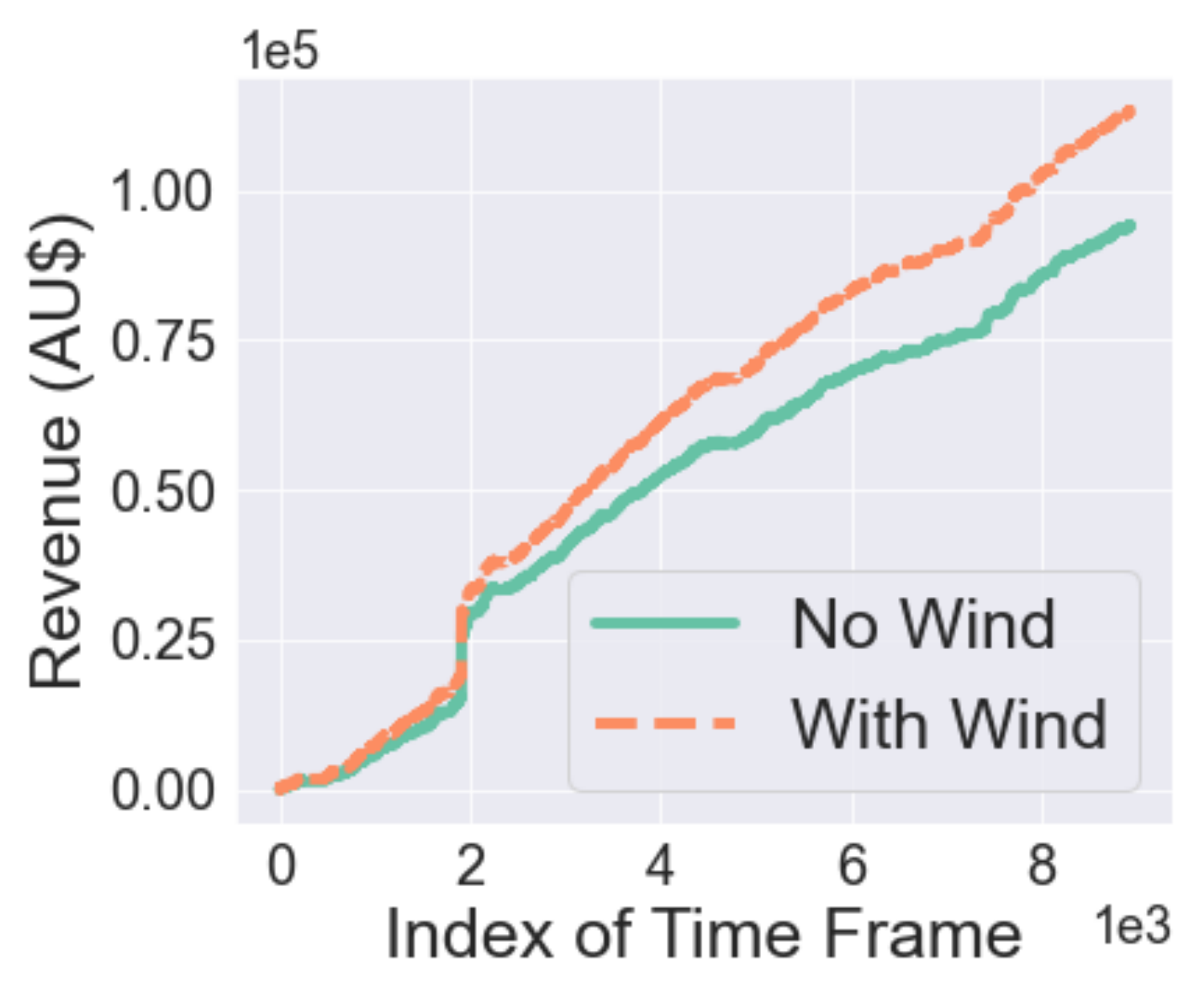}
    \label{fig:revenue_BESS_wwo_WC}
    }
    \hspace{-1em}
    \subfloat[Impacts of spot price and curtail frequency.]{
    \includegraphics[width=0.54\linewidth]{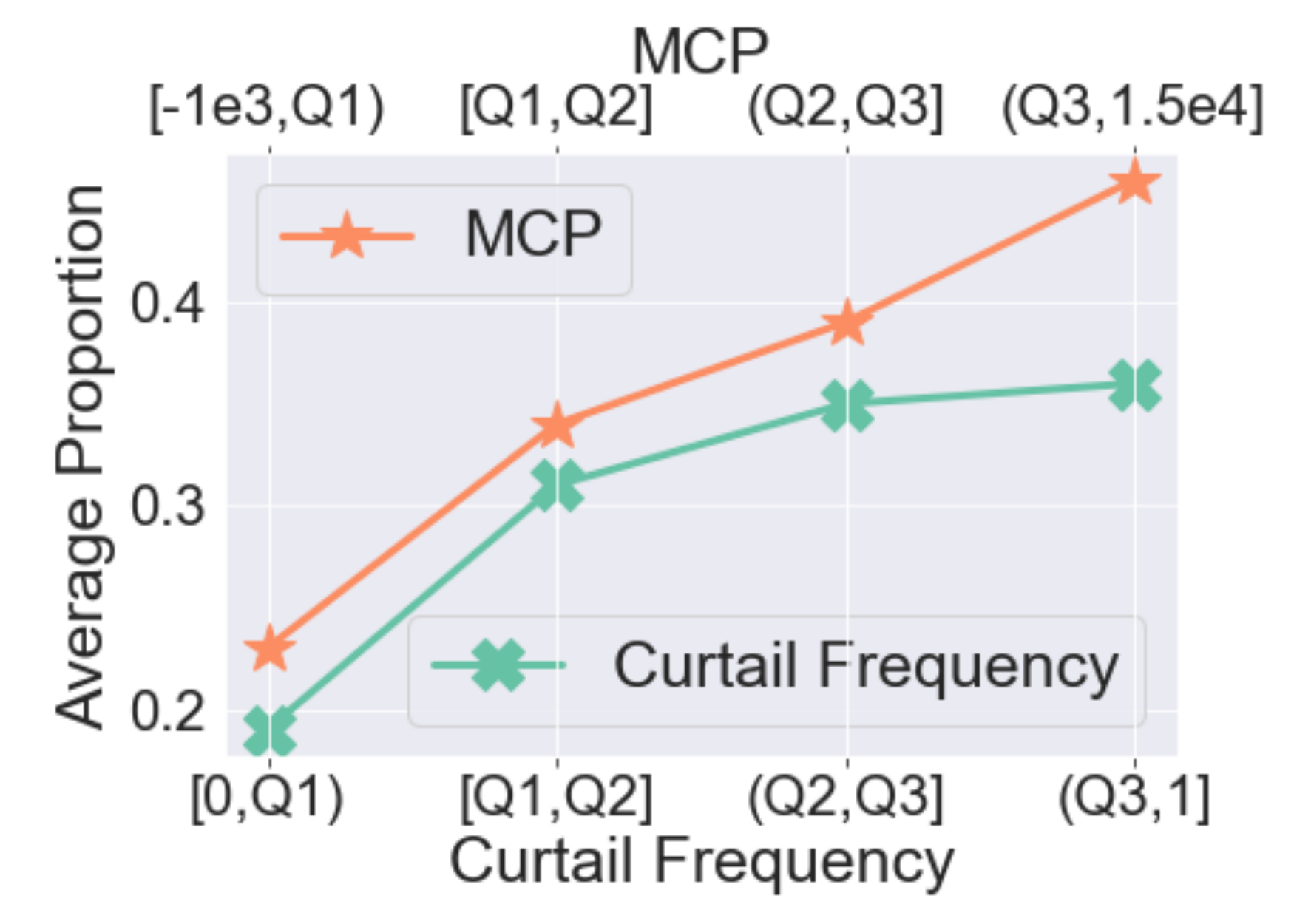}
    \label{fig:reserved_capacity4_WPC}
    }
    \caption{The effectiveness of wind curtailment reduction.}
    \label{fig:effectiveness_WC_redution}
    \vspace{-1.7em}
\end{figure}

According to Fig. \ref{fig:baseline_comparison_power_source}, the JointDRL chooses the curtailed wind as an important power source to charge the BESS for higher financial rewards. We examine the associated economic benefits by comparing the BESS bidding simulation results with/without using onsite curtailed wind energy. The results are illustrated in Fig. \ref{fig:revenue_BESS_wwo_WC}. Interestingly, using curtailed wind energy can boost the overall revenue of the BESS by about $20\%$. Since purchasing power from the spot market would lead to a revenue loss (when the spot price is non-negative), using curtailed wind energy seems to be more economical and subsequently improves the overall revenue.

Since the BESS can act differently subject to market conditions and the availability of onsite wind curtailment, we further examine the relationship between spot prices and the amount of energy drawn from wind curtailment to capture the most influencing factors behind the action. The result is shown in Fig. \ref{fig:reserved_capacity4_WPC}, where we group the spot prices using its quartiles labelled by $Q1_\rho,Q2_\rho,Q3_\rho$. The BESS follows the arbitrage guideline and buys power mostly from the spot market when prices are low. During periods of higher spot prices, the BESS favors curtailed wind energy, as charging at high spot prices is likely to result in significant financial losses. Using curtailed energy is free, but subject to availability.

BESS charging decisions are also heavily influenced by wind curtailment frequency. We investigate the BESS operational dynamics at different levels of wind curtailment frequency that are grouped via its quartiles $Q1_f,Q2_f,Q3_f$. At the lowest level of curtailment frequency, the BESS draws approximately $19\%$ of curtailed wind power. When curtailment occurs more frequently, the BESS charges more from the onsite wind farm to reduce curtailment until reaching a plateau at approximately $36\%$, as shown in Fig. \ref{fig:reserved_capacity4_WPC}. This leveling-off differs from that observed at high spot prices, where the BESS continues to purchase more than $50\%$ of power from the spot market to charge the BESS. This is largely driven by wind curtailment uncertainty because wind curtailment may not occur even at a higher likelihood.

\section{Conclusion} \label{sec:conclusions}
This paper highlights the importance of coordinated efforts between the wind farm and the BESS in co-location to improve their bidding performance in the electricity spot market. We developed a model-free DRL-based real-time bidding strategy to explore the full potential of the wind-battery system in joint bidding. The DRL algorithm seeks to maximize financial rewards by balancing BESS energy arbitrage with wind curtailment. Simulation results show that our proposed strategy outperforms the P\&O benchmark in terms of faster execution time and better financial performance for both the wind farm and the BESS. We further investigate the operational dynamics of the BESS under various wind curtailment frequencies and market conditions, leading to two interesting insights: 1) the onsite otherwise curtailed wind power is an effective source to charge the BESS for additional financial returns; 2) The BESS tends to use more curtailed wind energy when the spot price and wind curtailment frequency increase. Successful application of our proposed strategy could promote the co-location of renewable generation and storage assets, strengthening government policies for broader system benefits. 

\bibliographystyle{IEEEtran}
\bibliography{IEEEabrv}

\begin{thebibliography}{10}
\providecommand{\url}[1]{#1}
\csname url@samestyle\endcsname
\providecommand{\newblock}{\relax}
\providecommand{\bibinfo}[2]{#2}
\providecommand{\BIBentrySTDinterwordspacing}{\spaceskip=0pt\relax}
\providecommand{\BIBentryALTinterwordstretchfactor}{4}
\providecommand{\BIBentryALTinterwordspacing}{\spaceskip=\fontdimen2\font plus
\BIBentryALTinterwordstretchfactor\fontdimen3\font minus
  \fontdimen4\font\relax}
\providecommand{\BIBforeignlanguage}[2]{{%
\expandafter\ifx\csname l@#1\endcsname\relax
\typeout{** WARNING: IEEEtran.bst: No hyphenation pattern has been}%
\typeout{** loaded for the language `#1'. Using the pattern for}%
\typeout{** the default language instead.}%
\else
\language=\csname l@#1\endcsname
\fi
#2}}
\providecommand{\BIBdecl}{\relax}
\BIBdecl

\bibitem{CEC2022}
CEC, \emph{Clean Energy Australia Report}.\hskip 1em plus 0.5em minus
  0.4em\relax Clean Energy Council, 2022.

\bibitem{burke2011}
D.~J. Burke and M.~J. O'Malley, ``Factors influencing wind energy
  curtailment,'' \emph{IEEE Transactions on Sustainable Energy}, vol.~2, no.~2,
  pp. 185--193, 2011.

\bibitem{aemo_future_gen}
AEMO, \emph{{Australian National Electricity Market Generation
  Information}}.\hskip 1em plus 0.5em minus 0.4em\relax Australian Energy
  Market Operator, 2022.

\bibitem{AEMOSysPlan2022}
------, \emph{Integrated System Plan for the National Electricity
  Market.}\hskip 1em plus 0.5em minus 0.4em\relax {Australian Energy Market
  Operator}, 2022.

\bibitem{sun2017}
Y.~Sun, J.~Zhong, Z.~Li, W.~Tian, and M.~Shahidehpour, ``Stochastic scheduling
  of battery-based energy storage transportation system with the penetration of
  wind power,'' in \emph{2017 IEEE Power \& Energy Society General Meeting},
  2017, pp. 1--1.

\bibitem{alanazi2017}
A.~Alanazi and A.~Khodaei, ``Optimal battery energy storage sizing for reducing
  wind generation curtailment,'' in \emph{2017 IEEE Power \& Energy Society
  General Meeting}, 2017, pp. 1--5.

\bibitem{nikoobahkt2020}
A.~Nikoobakht, J.~Aghaei, M.~Shafie-Khah, and J.~P.~S. Catalão, ``Minimizing
  wind power curtailment using a continuous-time risk-based model of generating
  units and bulk energy storage,'' \emph{IEEE Transactions on Smart Grid},
  vol.~11, no.~6, pp. 4833--4846, 2020.

\bibitem{akbari2019}
E.~Akbari, R.-A. Hooshmand, M.~Gholipour, and M.~Parastegari, ``Stochastic
  programming-based optimal bidding of compressed air energy storage with wind
  and thermal generation units in energy and reserve markets,'' \emph{Energy},
  vol. 171, pp. 535--546, 2019.

\bibitem{khojasteh2021}
M.~Khojasteh, P.~Faria, and Z.~Vale, ``A robust model for aggregated bidding of
  energy storages and wind resources in the joint energy and reserve markets,''
  \emph{Energy}, vol. 238, p. 121735, 08 2021.

\bibitem{xie2021}
Y.~Xie, W.~Guo, Q.~Wu, and K.~Wang, ``Robust mpc-based bidding strategy for
  wind storage systems in real-time energy and regulation markets,''
  \emph{International Journal of Electrical Power \& Energy Systems}, vol. 124,
  p. 106361, 2021.

\bibitem{weron2014}
R.~Weron, ``Electricity price forecasting: A review of the state-of-the-art
  with a look into the future,'' \emph{International Journal of Forecasting},
  vol.~30, no.~4, pp. 1030--1081, 2014.

\bibitem{aemo2020}
AEMO, \emph{How the National Electricity Market works}.\hskip 1em plus 0.5em
  minus 0.4em\relax Australian Energy Market Operator, 2020.

\bibitem{aemo2022}
------, \emph{NEM Operational Forecasting and Dispatch Handbook for wind and
  solar generators}.\hskip 1em plus 0.5em minus 0.4em\relax Australian Energy
  Market Operator, 2022.

\bibitem{bordin2017}
C.~Bordin, H.~O. Anuta, A.~Crossland, I.~L. Gutierrez, C.~J. Dent, and D.~Vigo,
  ``A linear programming approach for battery degradation analysis and
  optimization in offgrid power systems with solar energy integration,''
  \emph{Renewable Energy}, vol. 101, pp. 417--430, 2017.

\bibitem{anwar2022}
M.~Anwar, C.~Wang, F.~de~Nijs, and H.~Wang, ``Proximal policy optimization
  based reinforcement learning for joint bidding in energy and frequency
  regulation markets,'' \emph{IEEE Power \& Energy Society General Meeting
  (PESGM)}, 2022.

\bibitem{fujimoto2018}
S.~Fujimoto, H.~Hoof, and D.~Meger, ``Addressing function approximation error
  in actor-critic methods,'' in \emph{International Conference on Machine
  Learning}, 2018, pp. 1582--1591.

\bibitem{aemo_data}
AEMO, \emph{AEMO Nemweb data}.\hskip 1em plus 0.5em minus 0.4em\relax
  Australian Energy Market Operator, 2022.

\bibitem{mitchell2011}
S.~Mitchell, M.~OSullivan, and I.~Dunning, \emph{PuLP: a linear programming
  toolkit for python}.\hskip 1em plus 0.5em minus 0.4em\relax The University of
  Auckland, 2011.

\end{thebibliography}
\end{document}